\documentclass[letterpaper, 10 pt, conference]{ieeeconf}  

\IEEEoverridecommandlockouts                              

\usepackage{graphicx}

\title{\LARGE \bf
Teaching Perception
}

\author{Jonathan H. Connell$^{1}$
\thanks{$^{1}$Jonathan Connell is with IBM T.J. Watson Research, Yorktown Heights NY, USA
        {\tt\small jconnell@us.ibm.com}}%
}

\begin{document}

\maketitle
\thispagestyle{empty}
\pagestyle{empty}

\begin{abstract}
The visual world is very rich and generally too complex to perceive in its entirety. Yet only certain features are typically required to adequately perform some task in a given situation. Rather than hardwire-in decisions about when and what to sense, this paper describes a robotic system whose behavioral policy can be set by verbal instructions it receives. These capabilities are demonstrated in an associated video \cite{video-p} showing the fully implemented system guiding the perception of a physical robot in simple scenario. The structure and functioning of the underlying natural language based symbolic reasoning system is also discussed.
\end{abstract}

\section{INTRODUCTION}

Sensing is not without costs. For any given object there are many things that can be known about it. What constitutes a reasonable amount of information to obtain? For instance, to identify an object in a scene a robot could run a DNN recognizer. But, depending on the resources available, this may take a noticeable amount of time. And, while some recognizers have N-ary outputs, others are designed as one-versus-all. In this case, to classify an object a robot might have to run N separate nets. But classification is not the end of the story, there are also properties like color, shape, and pattern. Does it make sense to run all these potential nets on all objects all the time? 

In biology there seems to be a division between the ``where" system and the ``what" system \cite{dorsal-ventral}. That is, an agent is either trying to find something in particular, or has already focused on an item and is trying to identify it. Generally, animals do not seem to try to create a complete description of a scene by finding and identifying every object present. This is one heuristic to limit the load imposed by perception.

Even when an object has already been picked out from the background, humans use a selective version of the ``what'' system. For instance, to identify a bird it is important to pay attention to field marks, such as the color of its legs and whether it has an eye ring. The bird's overall size and color are often less informative. Similarly, to determine the model year of a car one needs to examine the details of the front grill and the shape of the taillights. The obvious gross features of having four wheels and a number of windows are largely irrelevant.
Moreover, the same object may be viewed in different ways depending on the task at hand. When apprehending bank robbers, the police might instead focus on the face of the driver and the license plate of the getaway car.

The visual information described so far is largely overtly evident and can often be obtained relatively quickly. However, there are cases where an object needs to be manipulated or a new viewpoint sought in order to discover additional information. For example, when a robot is told ``Hand me that bottle" a simple shape predicate is all that is needed. Yet when asked to ``Hand me the Advil" it might need to see that backside of the bottle in order to read the label. Thus, obtaining this information is even more expensive. There have been a number of strategies in computer vision systems including animate vision \cite{Ballard}, purposive vision \cite{Aloimonos}, and active vision \cite{Bajcsy} that try to handle (and synergistically exploit) this interplay between perception and action.

Assessing other predicates can take even more work, such as deciding whether ``Mary likes cake". The robot might have to gather ingredients, mix the batter and cook it, and finally serve it to Mary to see if she smiles. Of course another option would be to simply ask Mary ``Do you like cake?", although this itself takes some time and Mary's response might not be truthful. Yet, even this might not be appropriate (or feasible) in all cases, such as deciding whether the Queen of England likes cake.

Some properties have more serious problems connected with them. 
A number of predicates require interaction with an object, such as lifting it to estimate its weight or density. This might be fine for items like books, but a bad idea for items like snakes. Similarly, to see if something is sleeping the robot could poke it. This is (sometimes) okay for people, but ill-advised for bears. Finally, consider ``flammable". This can be tested by attempting to light an item on fire. Yet, while this experiment might provide an answer, it can lead to an irreversible state change in which the robot is holding, say, charred shreds instead of the envelope containing a bill that needs to be paid. 

To gather complete information on any one object can potentially take an unlimited amount of time and is fraught with many perils. Deciding how much effort to invest and when certain exploratory procedures are appropriate depends on numerous criteria, some physical, some time-based, and some social. It is clear that a line has to be drawn somewhere, but how? While there have been attempts \cite{cost-sensitive} to design numeric cost-benefit analysis schemes to cover all these cases, it is relatively simple to explain to another human what to do and when. This is the approach taken with the robot described here. It has a language interpreter and associated reasoning engine that makes it capable of receiving and then acting on advice (perceptual and otherwise) supplied by its user. 

\begin{figure*}[t]
\centering
\includegraphics[width=0.9\textwidth]{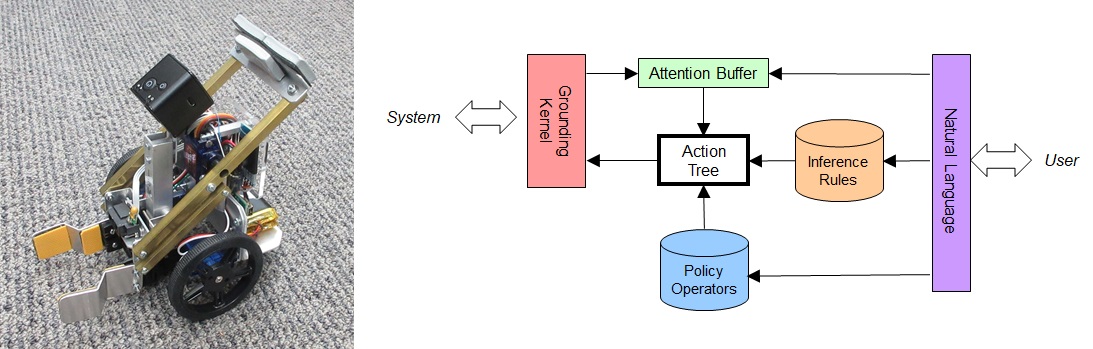} 
\caption{The ALIA reasoning system uses a set of rules to interpret situations, and a set of operators to generate behaviors. The output of the controller relies on a set of grounding functions intrinsic to the deployment platform, in this case a small forklift robot.}
\label{fig:block}
\end{figure*}

\section{RELATED WORK}

Aside from the early work on active perception (summarized in \cite{revisit}), there has been a lot of recent research on planning visitation locations for environmental mapping \cite{multi-map,ship-vSLAM}. These systems generally approach the problem as search with integrated field-of-view and movement constraints. There has also been work done on where to point a camera in order to best reconstruct a scene given a limited number of snapshots \cite{look-around}. In this case the problem was formulated as a reinforcement learning exercise. Similar research has been done for 3D modeling of objects from a series of next best views \cite{3D-viewpoints}. Note that in all these systems, the view selection is based on automatic optimization methods, not human-supplied advice.
 
Other work has been done with exploratory actions in manipulation \cite{push-poke}. Here to determine if an object was, say, metallic, the system would automatically select the action ``poke'' and listen for a characteristic sound, then ``look'' at the object's appearance, and finally choose to ``report'' its conclusion. The system couched the problem as a MOMDP with temporal costs for each action and a reward for the correctness of the result. While such time costs are straightforward, social costs (e.g. for ``poke'') are more difficult to estimate and not necessarily commensurate with time, making them difficult to incorporate in such frameworks.

There are also a number of robot systems that do some learning from language. For instance, the programming by demonstration system in \cite{verbal-prog} allows the user to invoke sophisticated motion commands via speech. The DIARC system \cite{DIARC-follow} can additionally be taught action expansions, such as what ``follow me'' means. Other systems \cite{Eli-AGI,DIARC-learn} can learn to associate naming terms with objects or sequential procedures. 
However none of these systems attempt to control perceptual processing through explicit linguistic instruction. 

Given the complexities of knowing when and how deeply to process sensory inputs, the ALIA architecture described in the next section provides a relatively simple mechanism  for supplying the necessary guidance in a manner suited to non-technical users.

\section{ALIA COGNITIVE ARCHITECTURE}
\label{sec:design}

We are endeavoring to build an Advice Taker \cite{McCarthy}, where the top level behavior of an agent can be changed by simply talking to it. This enables one-shot learning by essentially just remembering (with appropriate generalization). This is in contrast to systems based on statistical machine learning, which may require 100's of training examples to catch on. Quick learning is essential for one-off tasks where training effort cannot be amortized over a long service life. It is also useful for rapid customization (e.g., of robots) by end users. 

Our emphasis on acquiring procedural knowledge directly from natural language is what sets ALIA apart from other cognitive architectures.
As a consequence of this, we have chosen a symbolic rule-based reasoner as the core of our system. 
Language most conveniently reduces to symbolic constructs, while reasoning and action recommendations are most often communicated in small chunks like rules. 

Fig.~\ref{fig:block} shows the overall structure of the implemented robot controller. The reasoning subsystem is split into declarative and procedural parts, amplifying the dichotomy proposed in \cite{ACT-proc}. That is, there are inference rules that derive new facts, and separate policy operators that control action. Both of these can be taught directly by the user through the natural language interface. In operation, something akin to a goal is generally posted by the user to the attention buffer. A small amount of rule-based reasoning is done to elaborate the situation, then a policy operator is selected for execution in the action tree. To actually accomplish a task, the system relies on a grounding kernel that interfaces with the robot hardware.

\begin{figure}[t]
\centering
\includegraphics[width=0.95\columnwidth]{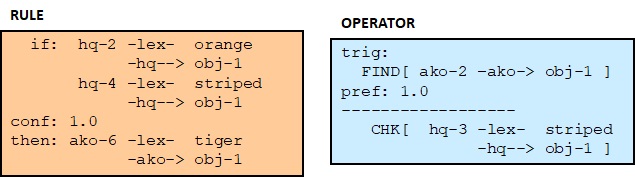}
\caption{Rules (left) are deductions based on working memory with conclusions asserted in the halo. Operators (right) are triggered by elements of the attention buffer and have a response composed of action directives.} 
\label{fig:rule-op}
\end{figure}

As shown in Fig.~\ref{fig:rule-op}, within the reasoning subsystem inference about facts is performed by rules (left), whereas advice about actions is conveyed by operators (right). 
Each rule can have a degree of belief in its conclusion, while each operator can have a preference value associated with its selection. 
The examples here correspond to the natural language inputs ``Orange striped things are (usually) tigers'' and ``To find out what something is, (you could) check if it is striped'', where the parenthesized elements correspond to belief and preference, respectively. 

The description of applicable situations is encoded as a semantic network, as are the details of the conclusions to be drawn or actions to be taken.
This declarative representation uses nodes for both objects (\texttt{obj-1}) and predicates (\texttt{hq-2}). The actual names assigned to the nodes are irrelevant and are merely for debugging purposes. Predicate nodes have directed links marked by the role for each argument. Here \textit{hq} means ``has quality'' and \textit{ako} means ``a kind of''. Each node can have a belief value
associated with it as well as a lexical item (such as ``orange'' for \texttt{hq-2}) giving the system a strong Whorfian flavor. 

These networks are relatively simple, along the lines of \cite{Boris-NL} rather than more elaborate formalisms such as \cite{SNePS,AMR}. The overall system is largely concerned with understanding procedural sequences and interpreting imperative commands. There is no explicit model of time and hence no tense or modality annotations are required, unlike in narrative comprehension. Moreover, our networks are built on-demand and disposed of a short while later. They do not accrete into a large knowledge base over time that needs to be updated and checked for consistency. 

\begin{figure}[t]
\centering
\includegraphics[width=0.95\columnwidth]{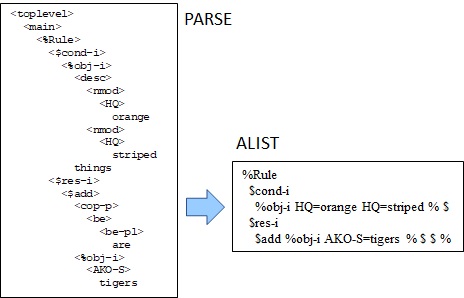}
\caption{The speech utterance \textit{Orange striped things are tigers} is first parsed by a semantic grammar then digested into an association list from which the actual rule is made.} 
\label{fig:parse}
\end{figure}

During instruction, suitable structured outputs are automatically derived from either text or speech -- they do not need to be entered in the intermediate form depicted here. Fig.~\ref{fig:parse} shows the steps in this process. The robot first listens to the user's speech as constrained by a Context Free Grammar. The resulting parse is then digested into an association list consisting of a number of slots and values, along with constituent bracketing, by walking the tree and retaining selected nodes. Finally, the a-list gets converted to the internal representation of the rule (Fig.~\ref{fig:rule-op} left) which is added to the database. 

The inference engine relies on a three-level memory structure. At the center is the attention buffer which holds recent facts and commands and is the primary conscious guide for the agent's activities. This can hold multiple focal items simultaneously. Once an item has been ``handled'' it is deactivated but persists in the attention buffer for a short amount of time. Outside attention there is working memory holding all the ancillary assertions linked to current or recently deactivated items. Finally, the outermost layer is the \textit{halo} which contains all those facts that are on the tip of your tongue but which only surface if needed. They remain unconscious in the sense that they cannot be used to make further deductions. 

On each cycle we recompute \textit{all} implications of the facts currently in memory. However, we limit deduction to two steps because unbridled forward chaining can lead to runaway. Moreover, human-supplied rules are often only locally consistent (bounded rationality) and can lead to global contradictions. 
A version of concatenation \cite{MACROP} or chunking \cite{SOAR} can be used to get deeper lookahead, although this is often not necessary. Note that because the halo is conceptually ephemeral and is constantly being re-derived as the contents of working memory change, having a limited horizon saves substantial work.

\begin{figure*}[t]
\centering
\includegraphics[width=0.7\textwidth]{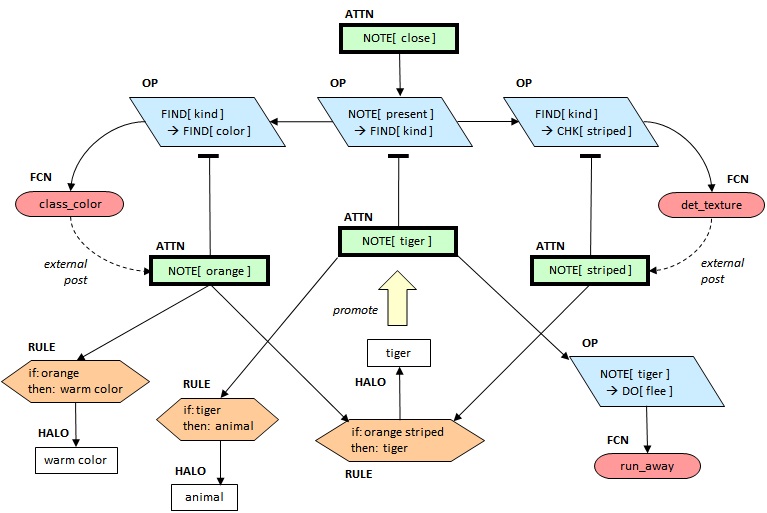} 
\caption{Attention items (green), operators (blue), and rules (orange) are combined to determine that the agent should run away from a tiger that has just appeared (top).}
\label{fig:tiger}
\end{figure*}

The action engine also works via forward chaining, where stimulus leads to response. This allows the system to spontaneously react to circumstances and exhibit initiative. Each of our operators has a trigger condition akin to a goal, and an enablement specification more like a typical antecedent. If the trigger is not matched to an item in the attention buffer, the operator is never considered for use. 
The invoked procedures consist of a number of \textit{directives}, intentional tags attached to particular nodes in the network. The full list includes: NOTE, DO, ANTE, POST, CHK, FIND, ACH, KEEP, PUNT, and FCN. In the natural language front-end verbs are typically mapped to DO, yes/no questions about predicates to CHK, and wh- questions about objects to FIND. Another important flavor of directive is FCN, which provides grounding for the system by activating an external processing system of some sort. ALIA is not intended to control all details of the agent, but instead provides a sort of glue to bind things together on-the-fly based on verbal input. 

Of course backwards chaining is also useful, particularly for responding to user-imposed goals. Yet language features, such as \texttt{cut} in Prolog, are often introduced to keep logic-based systems from needlessly chasing dead ends. Some systems separately authorize chaining direction on a rule-by-rule basis \cite{logic-FB}. Our approach is based loosely on \textit{determinations}~\cite{determine}. These are not exact implications but rather functional dependencies like ``if you know a person's nationality, you likely know the language he speaks''.
This piece of knowledge can be translated into the forward chaining advice ``if you want to know what language a person speaks, determine his nationality.'' A separate forward chaining set of rules would then capture the dependencies, such as ``if a person is from Brazil, he likely speaks Portuguese''. 
The operator and rule shown in Fig.~\ref{fig:rule-op} implement exactly this idea, which is reminiscent of the buffer filling then harvesting pattern found in EPIC \cite{EPIC}. 

The actual interplay between rules and operators to guide perception will be detailed in the next section. Note, however, that the ALIA system can also be used to instill procedures, reactions, prohibitions, and permissions via speech \cite{video}.

\section{Elaboration and Reaction}
\label{sec:tiger}

Let us illustrate the operation of the system using an example where an agent spontaneously decides to run away from a tiger. While this is an unlikely scenario for any robot, it does highlight several of the key methods in the architecture. In some sense the presence of the tiger provides the affordance of running away, but the system must be able to recognize this. Fig.~\ref{fig:tiger} details the actual process.

In the very first step the perception system posts an item to the attention buffer saying that a new visual object is near by. This is abbreviated as the green NOTE[close] box near the top of the diagram. The real semantic net assertion would be something like:
\texttt{obj-1~<-hq-~close}. 

Next the system invokes an operator (blue) encapsulating the advice that if something is close, the system should try to identify what it is. The operator is triggered by the matching the NOTE directive in attention and sets a new imperative focus comprising a FIND[kind] directive. FIND tries \textit{all} applicable operators, one at a time, until a match has been found in working memory to the description in the directive. Here there are two possible follow-on operators that might be triggered. Assuming their preferences are the same, the system randomly picks, say, the left one first.

This operator says that, to find out what something is, it is helpful to find what color the object is. The new FIND[color] directive is then matched directly to the system grounding function \textit{class\textunderscore{}color}, shown in red. This invokes another piece of the visual system, possibly a color histogramming component, with an area of interest set to the original object that appeared. If this function fails, another method for determining color will be sought by automatic backtracking. However, if the function succeeds, it does not return the value to its caller (the operator). Instead, \textit{class\textunderscore{}color} directly posts its answer as a new item in the attention buffer. Here it adds the focus abbreviated as NOTE[orange], shorthand for the graphlet: \texttt{obj-1~<-hq-~orange~<-ako-~color}.

The new attention item satisfies the precondition of a deductive rule (lower left) which makes a tentative assertion in the unconscious halo memory noting that orange is a warm color. This is not currently relevant, so no operators are invoked on the basis of this inference. However, because the attention item is an assertion about color, it causes the most proximate FIND[color] directive to declare success and thus deactivate itself. This lets the higher level FIND[kind] directive invoke whatever operator is next in its list, in this case the one on the right calling for CHK[striped] (cf. Fig.~\ref{fig:rule-op}).

This directive is directly linked to the invocation of grounding function \textit{det\textunderscore{}texture}. As before, this runs some arbitrarily complex piece of code, perhaps a DNN, and injects its answer directly into the attention buffer. In addition, there may be extra physical steps required, such as orienting a narrow angle camera toward the object of interest to ensure sufficient resolution. Assuming that the process does not fail (e.g., the head might be busy doing something else) the NOTE[striped] assertion will be posted, which causes the successful completion of the calling CHK operator.  

Starting with just the bare assertion that something new has appeared, the system now knows that the object is orange and striped. This conjunction of facts triggers the rule at the bottom of the diagram which raises the suspicion that the object is a tiger (cf. Fig.~\ref{fig:rule-op} again). Because the top-level FIND was trying to determine just such a category, the halo assertion is promoted to a conscious attention item to fulfill the required pattern. At this point, the top-level operator is considered successfully completed and becomes quiescent.

Yet the assertion NOTE[tiger] has several ramifications. First, it triggers a rule which creates a halo fact that the object is also an animal. But there are no operators waiting for this missing piece. On the other hand, the new assertion directly matches the trigger clause of the final operator (lower right) which energizes the directive DO[flee]. This in turn can be realized by the grounding function \textit{run\textunderscore{}away} to spur the system into action. Again, these system calls can be quite sophisticated and time consuming. For instance, running away might involve first orienting away from the stimulus, then engaging a cyclic pattern generator to move the legs while simultaneously activating the collision avoidance behavior to swerve around obstacles.

It is possible that one or more of these routines might fail, such as if the robot is driven into a corner. At this point it is up to \textit{run\textunderscore{}away} to signal failure so that its calling operator can look for some other method for DO[flee]. More likely, at some point \textit{run\textunderscore{}away} will signal successful completion because the reference object (tiger) is no longer visible, because the robot is sufficiently far away, or simply because the routine has timed out and is tired of running. When this occurs, no further actions are generated based on the initial stimulus.

\section{VIDEO DEMONSTRATION}

The video \cite{video-p} follows exactly the situation in Fig.~\ref{fig:tiger}. In this implementation the actual 
robot is fairly simple and the bulk of the processing (speech, reasoning, control) happens on a nearby Bluetooth-connected laptop. 
There are four degrees of freedom to control: two wheels, a gripper, and a lift stage. The associated grounding kernel responds to the commands drive, turn, grab/release, and raise/lower. Each of these is configured as a discrete instead of continuing action, although both modes could be present at the same time. That is, ``turn'' causes the robot to reorient by a fixed angle rather than spinning forever. The microcontroller on the robot takes care of generating pulses for the servos based on a serial communications line, but simple timeout-based action amounts are governed by the laptop. 
 For sensing, the robot relies primarily on a wifi-connected color camera. There is also a forward facing triangulation-based rangefinder mounted over the gripper,  but it is not used here.

The demonstration has the user talking to the nearby laptop, although a typing interface also exists. The program runs under Windows 10 and uses Microsoft Speech Recognizer 8.0 through the SAPI calls. Although this is a relatively old engine, it has the advantage of running locally so no network connection is required. Also, speech recognition still has some accuracy problems and so can benefit from whatever additional constraints can be imposed. This engine allows speaker-dependent acoustic models and can be run in a mixed grammar/dictation mode. The use of grammar can not only limit the vocabulary to reasonable terms, but can also enforce preferred phrasing patterns. Thus, when the system is told to ``pick up the block'' by a speaker with a foreign accent, it never erroneously hears ``peacock the block''. Although ``peacock'' is a valid English word, it makes no sense in the robot deployment context. Such mistakes can be difficult to detect and repair after the fact.

\begin{figure}[t]
\centering
\includegraphics[width=0.9\columnwidth]{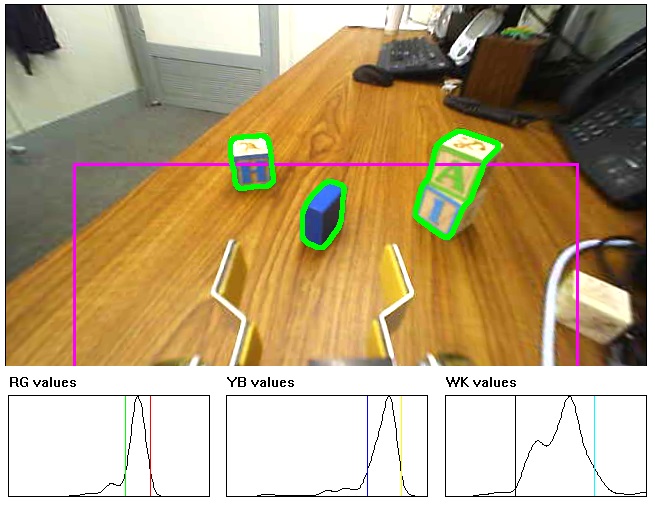} 
\caption{The magenta region (top) is used is used to find distributions in opponent color space (bottom). Objects (outlined in green) are detected as gaps in this presumed uniform surface color.}
\label{fig:detect}
\end{figure}

As stated, the vision system uses a wifi-linked color camera with a linear polarizer to suppress glare from shiny surfaces. The input image is corrected for lens distortion, contrast enhanced, then temporally smoothed. All of this is in pursuit of stable color, which is important for object segmentation and characterization. Because this robot is meant only as a technology demonstrator, its object detection system is fairly simple. As in \cite{Eli} the robot builds a model on-the-fly of the supporting surface and looks for deviations from this, as shown in Fig.~\ref{fig:detect}. This lets it detect objects it has never seen before, unlike other approaches \cite{YOLO,maskCNN} that require pre-trained models for all important objects. Gaps in the background are smoothed, filtered by criteria like rough shape and size, and then tracked over time. 

Detection and tracking are the only operations performed continuously. However the grounding kernel does automatically monitor the distance of each object, as determined by the height of its bottom in the image. When something enters the personal space of the robot (within about 10cm in front), the vision system spontaneously posts a message to the the ALIA attention buffer. This is what sets off the cascade of reasoning in Fig.~\ref{fig:tiger}. As a side effect, the node name used in the assertion is bound to the detected object. Tracking makes sure the label stays with this particular object for later reference.

As explained in the previous section, the architecture can deliberately query the grounding kernel about the color of some object using the \textit{class\textunderscore{}color} function. Converting from pixels to color terms is done in several steps. First, the portion of the image corresponding to the object is converted to HSI color space, then colorful pixels of the object are found by checking for sufficient saturation and single channel pixel values. Pixels that pass these tests are segregated into six different color bins by hue: red, orange, yellow, green, blue, and purple. Non-colored pixels are deemed black, gray, or white based on intensity. After each pixel has been given one of these nine canonical colors, all the pixels in the object are histogrammed. A final color determination of one or more values is based on the relative populations of these bins. The grounding kernel then adds one or more assertions (NOTEs) using the object's identifier plus the textual names of the colors found. 

ALIA can also ask that the grounding kernel check the visual texture of the object using the function \textit{det\textunderscore{}texture}. In particular, to judge whether the object is striped or not it first convolves the monochrome intensity image with separate horizontal and vertical Sobel filters. The responses are thresholded and then grouped by connected components. Only edges greater than some minimum length are retained and then the fraction of the foreground mask covered by such lines is estimated. For an object to be striped, there must be more than a certain number of lines and the area coverage must exceed a threshold. The grounding kernel then directly posts a NOTE to the attention buffer that the object with the given identifier either is, or is not, striped.

In the video the robot is commanded to drive forward slowly. At a certain point the tiger, which has been tracked for a while as a generic object, it deemed close enough to alert the reasoning system. This in turn activates the analysis routines for color and texture. Once the threat is identified, a response is chosen, namely to flee. Here, this is accomplished through an intermediate operator that was previously trained based on the verbal instruction:

\begin{center}
\textit{
To flee move backward and say save me master
}
\end{center}

\noindent There might conceivably be other operators for running away, such as turning to face the object while backing up, or uttering some other exclamation. If so, the architecture chooses probablistically between them based on the associated preferences.

\begin{figure}[t]
\centering
\includegraphics[width=0.9\columnwidth]{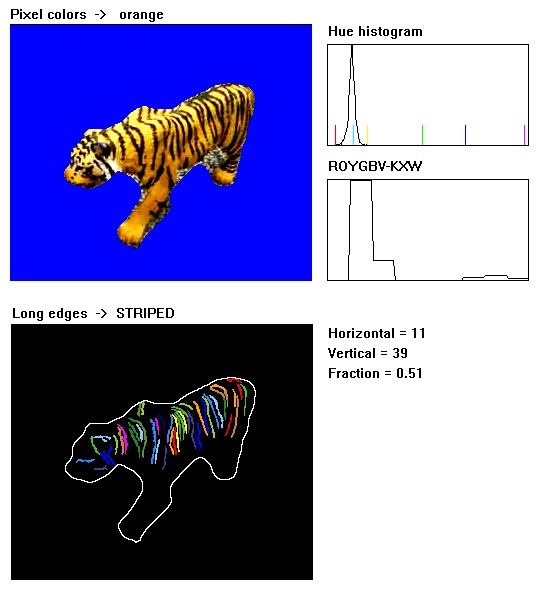} 
\caption{The color(s) associated with an object are found by canonicalizing each pixel (top) then histogramming. Stripedness is evaluated by determining the length and extent of Sobel edges found in the object mask (bottom).}
\label{fig:props}
\end{figure}

\section{CONCLUSION}
\label{sec:concl}

The example focuses on running away from a (toy) tiger. However the perception control is general enough that we can instill additional reactions using only a few sentences, such as below.

\begin{center}
\textit{
A black and white and striped thing is a zebra \\
If a zebra is close then stop and beep
}
\end{center}

\noindent This adds one additional halo rule and one operator, but now the robot will attempt to startle zebras out of its path. 

Contrast this arrangement with what would happen under the more specific set of perceptual instructions:

\begin{center}
\textit{
If something is close then find out what color it is \\
If something is orange check if it is striped
}
\end{center}

\noindent This would result in the same action in the case of the tiger, running away, but responding to the zebra would take more work than before because its stripes are not noticed. Both of these perceptual policies can be conveyed by language; they just represent different ways of looking at the world. In some sense the second policy is lazier -- it always checks the color of a nearby thing hoping that this will trigger some appropriate reaction, but only looks for stripes in certain cases. This is like the old search heuristic of horizon deepening used in chess when certain pivotal positions are encountered. 

In summary, we have described a symbolic rule-based cognitive architecture in which all assertions, commands, rules, and operators are entered via (spoken) natural language. The system is driven by an attention buffer supplemented by context from an associated working memory.  Outside of this there is a seething froth of inference that remains unconscious until certain elements are pulled in by the action system. As shown in the worked-out example (and video), the system can proactively respond to environmental conditions, even when physical or mental effort needs to be exerted to fully comprehend the situation.

\addtolength{\textheight}{-11.5cm}   


\bibliography{biblio}

\begin{thebibliography}{10}
\providecommand{\url}[1]{#1}
\csname url@rmstyle\endcsname
\providecommand{\newblock}{\relax}
\providecommand{\bibinfo}[2]{#2}
\providecommand\BIBentrySTDinterwordspacing{\spaceskip=0pt\relax}
\providecommand\BIBentryALTinterwordstretchfactor{4}
\providecommand\BIBentryALTinterwordspacing{\spaceskip=\fontdimen2\font plus
\BIBentryALTinterwordstretchfactor\fontdimen3\font minus
  \fontdimen4\font\relax}
\providecommand\BIBforeignlanguage[2]{{%
\expandafter\ifx\csname l@#1\endcsname\relax
\typeout{** WARNING: IEEEtran.bst: No hyphenation pattern has been}%
\typeout{** loaded for the language `#1'. Using the pattern for}%
\typeout{** the default language instead.}%
\else
\language=\csname l@#1\endcsname
\fi
#2}}

\bibitem{video-p}
\BIBentryALTinterwordspacing
J.~Connell, ``Teaching perception (video),'' 2019. [Online]. Available:
  \url{https://youtu.be/jZT1muSBjoc}
\BIBentrySTDinterwordspacing

\bibitem{dorsal-ventral}
M.~Goodale and D.~Milner, ``Separate visual pathways for perception and
  action,'' \emph{Trends in Neuroscience}, vol.~15, no.~1, pp. 20--–25, 1992.

\bibitem{Ballard}
D.~H. Ballard, ``Animate vision,'' \emph{Artificial Intelligence}, vol.~48, pp.
  57–--86, 1991.

\bibitem{Aloimonos}
J.~Y. Aloimonos, ``Purposive and qualitative active vision,'' in
  \emph{International Conference on Pattern Recognition}, 1990, pp. 346--360.

\bibitem{Bajcsy}
R.~Bajcsy, ``Active perception,'' \emph{Proceedings of the IEEE}, vol.~76,
  no.~8, pp. 966–--1005, 1988.

\bibitem{cost-sensitive}
M.~Tan, ``Cost-sensitive learning of classification knowledge and its
  applications in robotics,'' \emph{Machine Learning}, vol.~13, no.~1, pp.
  7–--33, 1993.

\bibitem{revisit}
R.~Bajcsy, Y.~Aloimonos, and J.~Tsotsos, ``Revisiting active perception,''
  \emph{Autonomous Robots}, vol.~42, pp. 177–--196, 2018.

\bibitem{multi-map}
G.~Best, J.~Faigl, and R.~Fitch, ``Multi-robot path planning for budgeted
  active perception with self-organising maps,'' in \emph{Proc. IEEE Int. Conf.
  on Intelligent Robots and Systems (IROS)}, 2016, pp. 3164--3171.

\bibitem{ship-vSLAM}
A.~Kim and R.~Eustice, ``Perception-driven navigation: Active visual slam for
  robotic area coverage,'' in \emph{Proc. IEEE Int. Conf. on Robotics and
  Automation (ICRA)}, 2013, pp. 3196--3203.

\bibitem{look-around}
S.~Ramakrishnan, D.~Jayaraman, and K.~Grauman, ``Emergence of exploratory
  look-around behaviors through active observation completion,'' \emph{Science
  Robotics}, vol.~4, no.~30, 2019.

\bibitem{3D-viewpoints}
S.~Kriegel, M.~Brucker, Z.~Marton, T.~Bodenmuller, and M.~Suppa, ``Combining
  object modeling and recognition for active scene exploration,'' in
  \emph{Proc. IEEE Int. Conf. on Intelligent Robots and Systems (IROS)}, 2013,
  pp. 2384--2391.

\bibitem{push-poke}
S.~Amiri, S.~Wei, S.~Zhang, J.~Sinapov, J.~Thomason, and P.~Stone,
  ``Multi-modal predicate identification using dynamically learned robot
  controllers,'' in \emph{Proc. Int. Joint Conf. on Artificial Intelligence
  (IJCAI)}, 2018, pp. 4639--4645.

\bibitem{verbal-prog}
M.~Forbes, R.~Rao, L.~Zettlemoyer, and M.~Cakmak, ``Robot programming by
  demonstration with situated spatial language understanding,'' in \emph{Proc.
  IEEE Int. Conf. on Robotics and Automation (ICRA)}, 2015, pp. 2014--2020.

\bibitem{DIARC-follow}
R.~Cantrell, P.~Schermerhorn, and M.~Scheutz, ``Learning actions from
  human-robot dialogues,'' in \emph{IEEE Int. Symp. on Robot and Human
  Interactive Communication (RO-MAN)}, 2011, pp. 125--130.

\bibitem{Eli-AGI}
J.~Connell, E.~Marcheret, S.~Pankanti, M.~Kudoh, and R.~Nishiyama, ``An
  extensible language interface for robot manipulation,'' in \emph{Proc. of
  Conf. on Artificial General Intelligence (AGI)}, 2012, pp. 21--30.

\bibitem{DIARC-learn}
M.~Scheutz, E.~Krause, B.~Oosterveld, T.~Frasca, and R.~Platt, ``Spoken
  instruction-based one-shot object and action learning in a cognitive robotic
  architecture,'' in \emph{Proc. of Int. Conf. on Autonomous Agents and
  Multiagent Systems (AA-MAS)}, 2017, pp. 1378--1386.

\bibitem{McCarthy}
J.~McCarthy, ``Programs with common sense,'' in \emph{Symposium on
  Mechanization of Thought Processes}.\hskip 1em plus 0.5em minus 0.4em\relax
  National Physical Laboratory, UK, 1958, pp. 1--10.

\bibitem{ACT-proc}
J.~Anderson, \emph{Rules of the Mind (Ch. 2)}.\hskip 1em plus 0.5em minus
  0.4em\relax Lawrence Erlbaum, 1993.

\bibitem{Boris-NL}
B.~Katz, ``A three-step procedure for language generation,'' MIT, AI Memo
  AIM-599, 1980.

\bibitem{SNePS}
S.~Shapiro and W.~Rappaport, ``An introduction to a computational reader of
  narrative,'' in \emph{Deixis in Narrative: A Cognitive Science Perspective},
  J.~Duchan \emph{et~al.}, Eds.\hskip 1em plus 0.5em minus 0.4em\relax Lawrence
  Erlbaum, 1995, pp. 79--105.

\bibitem{AMR}
C.~Banarescu \emph{et~al.}, ``Abstract meaning representation for sembanking,''
  in \emph{Proc. Linguistic Annotation Workshop}.\hskip 1em plus 0.5em minus
  0.4em\relax Assoc. for Comp. Linguistics, 2013, pp. 178--186.

\bibitem{MACROP}
R.~Fikes and N.~Nilsson, ``Learning and executing generalized robot plans,''
  \emph{Artificial Intelligence}, vol.~3, no.~4, pp. 251--288, 1972.

\bibitem{SOAR}
J.~Laird, P.~Rosenbloom, and A.~Newell, ``The anatomy of a general learning
  mechanism,'' \emph{Machine Learning}, vol.~1, no.~1, pp. 11--46, 1986.

\bibitem{logic-FB}
\BIBentryALTinterwordspacing
R.~Kowalski and F.~Sadri, ``Introduction to logic-based production systems,''
  2016. [Online]. Available:
  \url{https://www.doc.ic.ac.uk/\textasciitilde{}rak/papers/LPS with CLOUT.pdf}
\BIBentrySTDinterwordspacing

\bibitem{determine}
T.~Davies and S.~Russell, ``A logical approach to reasoning by analogy,'' in
  \emph{Proc. IJCAI-87}.\hskip 1em plus 0.5em minus 0.4em\relax Morgan
  Kaufmann, 1987, pp. 264--270.

\bibitem{EPIC}
D.~Kieras and D.~Meyer, ``An overview of the epic architecture for cognition
  and performance with application to human-computer interaction,''
  \emph{Human-Computer Interaction}, vol.~12, no.~4, pp. 391--438, 1997.

\bibitem{video}
\BIBentryALTinterwordspacing
J.~Connell, ``Teachable systems (video),'' 2019. [Online]. Available:
  \url{https://youtu.be/EjzdjWy3SKM}
\BIBentrySTDinterwordspacing

\bibitem{Eli}
\BIBentryALTinterwordspacing
------, ``Extensible grounding of speech for robot instruction,'' in
  \emph{Robots That Talk and Listen}, J.~Markowitz, Ed.\hskip 1em plus 0.5em
  minus 0.4em\relax De Gruyter, 2015, pp. 175--202. [Online]. Available:
  \url{https://arxiv.org/abs/1807.11838}
\BIBentrySTDinterwordspacing

\bibitem{YOLO}
\BIBentryALTinterwordspacing
J.~Redmon and A.~Farhadi, ``Yolov3: An incremental improvement,'' 2018.
  [Online]. Available: \url{https://arxiv.org/abs/1804.02767}
\BIBentrySTDinterwordspacing

\bibitem{maskCNN}
K.~He, G.~Gkioxari, P.~Dollar, and R.~Girshick, ``Mask r-cnn,'' in \emph{Proc.
  IEEE Int. Conf. on Computer Vision (ICCV)}, 2017, pp. 2980--2988.

\end{thebibliography}
\bibliographystyle{IEEEtran}

\end{document}